\documentclass{hld2024}

\usepackage{booktabs}

\title[Progress Measures for Grokking on Real-world Tasks]{Progress Measures for Grokking on Real-world Tasks}

\hldauthor{%
\Name{Satvik Golechha} \Email{zsatvik@gmail.com}\\
\addr {Independent Researcher} 
}

\begin{document}

\maketitle

\begin{abstract}%

Grokking, a phenomenon where machine learning models generalize long after overfitting, has been primarily observed and studied in algorithmic tasks. This paper explores grokking in real-world datasets using deep neural networks for classification under the cross-entropy loss. We challenge the prevalent hypothesis that the $L_2$ norm of weights is the primary cause of grokking by demonstrating that grokking can occur outside the expected range of weight norms. To better understand grokking, we introduce three new progress measures: activation sparsity, absolute weight entropy, and approximate local circuit complexity. These measures are conceptually related to generalization and demonstrate a stronger correlation with grokking in real-world datasets compared to weight norms. Our findings suggest that while weight norms might usually correlate with grokking and our progress measures, they are not causative, and our proposed measures provide a better understanding of the dynamics of grokking.

\end{abstract}

%\begin{keywords}%
%  List of keywords%
%\end{keywords}

\section{Introduction}

Machine learning models are typically trained and evaluated using different datasets. Grokking \citep{power2022grokking} is a phenomenon where models generalize to low test-loss long after they overfit their training data. Grokking was first observed in algorithmic tasks \cite{power2022grokking}, and several works  have since attempted to understand why and how grokking happens in these algorithmic tasks \citep{liu2022towards, nanda2023progress, varma2023explaining}. 

\citet{liu2022omnigrok} show that grokking can be observed in a ``goldilocks zone'' of weight norms on deep networks trained on non-algorithmic tasks. We follow their setup and consider the simple problem of multi-class classification using neural networks trained with the cross-entropy loss. The MNIST dataset \cite{deng2012mnist} contains image inputs $X$, a set of labels $Y$, and a training dataset, $D = \{(x_1, y_{1}), (x_2, y_{2}), \ldots, (x_d, y_{d})\}$.

For any arbitrary model acting as a classifier $h : X \times Y \rightarrow \mathbb{R}$, the softmax cross-entropy loss is given by:

$$
L_{\text{CE}}(h) = -\frac{1}{D} \sum_{(x, y) \in D} \log \frac{\exp(h(x, y))}{\sum_{y' \in Y} \exp(h(x, y'))}.
$$

The output of our model are the logits $o_h(x)$, which can give the class-wise output probabilities as $\text{softmax}(o^y_h(x)) \triangleq h(x, y)$. The set of all the parameters of the model (typically a neural network) is denoted by $\theta$ and the $L_2$ norm of the weights is denoted by $\|\theta\|^2_2$.

Our main contributions are as follows:

\begin{itemize}
    \item We contradict the hypothesis of \citet{liu2022omnigrok} that weight norms explain grokking by showing that generalization can occur way outside the ``goldilocks zone'' of low weight norm.
    \item We conceptually motivate three alternative progress measures for grokking on real-world data and show that they align well with generalization on a simple neural network trained on MNIST \citep{deng2012mnist} and an LSTM-based model trained on the IMDb dataset \citep{maas2011learning}.
\end{itemize}

\section{Related Work}

Grokking has been observed in both algorithmic \citep{liu2022omnigrok, power2022grokking} and real-world tasks \citep{humayun2024deep}. Several attempts have been made to explain grokking \citep{liu2022towards, nanda2023progress, varma2023explaining} in algorithmic tasks. Real-world data and models have been explored much less in terms of grokking, with \citet{humayun2024deep, liu2022omnigrok} showing the phenomenon to occur in the first place, and attempting to explain them using some heuristics. 

Progress measures \citep{barak2022hidden} have been a powerful tool to understand deep networks by analyzing their training dynamics. \citet{nanda2023progress, varma2023explaining, liu2022omnigrok} introduce various measures to explain grokking on mathematical tasks, and \citet{liu2022omnigrok} introduce the $L_2$ weight norm as the cause of grokking on real-world tasks. We show that this prevalent metric does not explain grokking in real-world tasks, and introduce three conceptually motivated progress measures that do.

\section{$L_2$ Norms of Weights do not Explain Grokking}

Unlike algorithmic tasks, for which grokking has been observed and studied in detail \citep{nanda2023progress, liu2022towards, varma2023explaining}, there has been limited progress made to elicit and understand grokking in neural network models trained on real-world datasets. This was first observed by \citet{liu2022omnigrok}, where they found that by training a model using weight decay and starting with a high weight initialization, grokking can be observed on real-world datasets (such as MNIST \citep{deng2012mnist}).

\begin{figure}[h!]
\centering
\includegraphics[width=\textwidth, trim=1.5cm 1.5cm 1.5cm 1cm, clip]{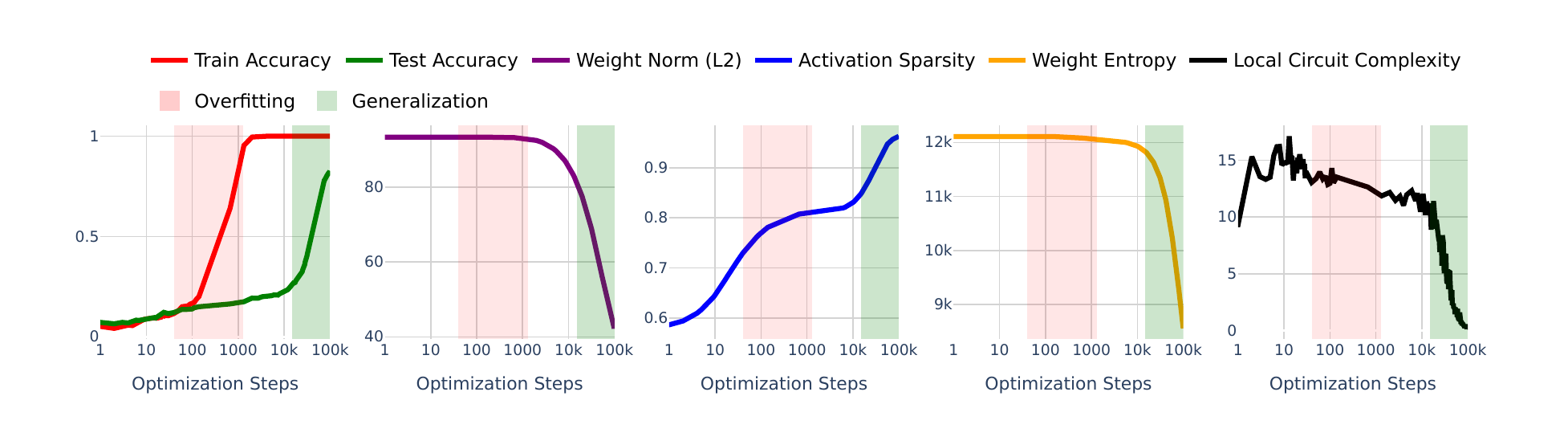}
\caption{Grokking, as observed on MNIST by \citet{liu2022omnigrok}, with a reduction in the $L_2$ norm of the weights and the new progress measures introduced.}
\label{fig:decreasing_norm_mnist}
\end{figure}

\begin{figure}[h!]
\centering
\includegraphics[width=\textwidth, trim=1.5cm 1.5cm 1.5cm 1cm, clip]{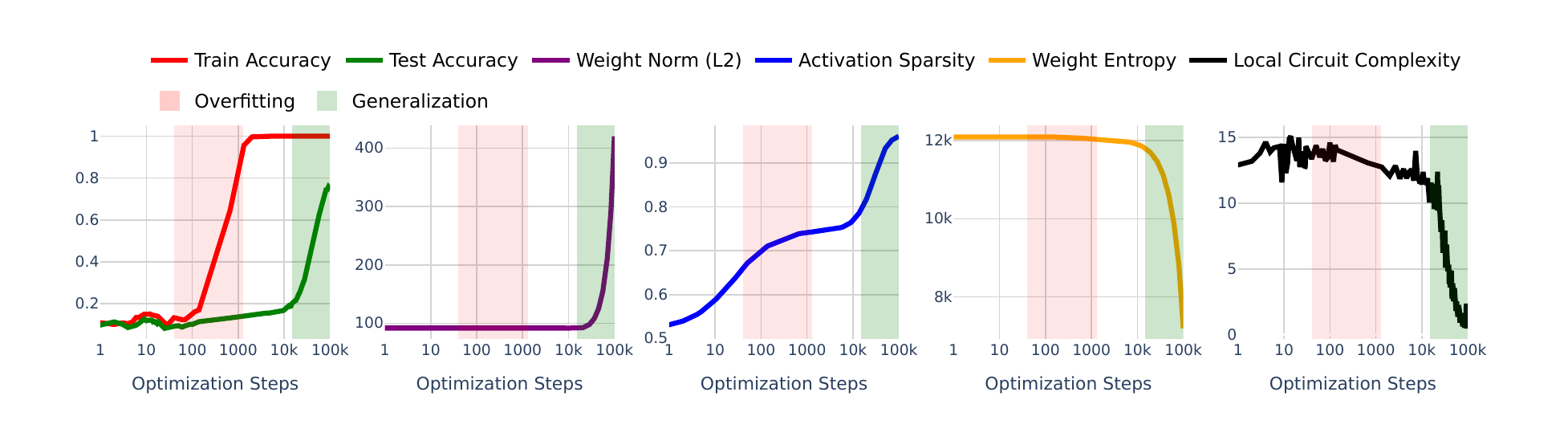}
\caption{Loss landscape for the modified setup showing an increase in the $L_2$ norm of the weights and the new progress measures introduced.}
\label{fig:increasing_norm_mnist}
\end{figure}

They hypothesized the weight norm to be the cause of grokking, which can occur in a range they call the "goldilocks zone". In this section, we refute this claim by showing that grokking can occur way outside this "goldilocks zone".

We employ a clever trick to elicit grokking while the $L_2$ norm of the weights increases during generalization. We keep the setup the same as in \citet{liu2022omnigrok}, and train a $3$-layer, $200-width$ MLP with ReLU non-linearities on the MNIST dataset using AdamW optimizer with weight decay. The hyperparameters are given in Appendix \ref{hyperparams_mnist}. We see grokking and a reduction in the norm of the weights in the default setup. (see Fig. \ref{fig:decreasing_norm_mnist}).

We then add the squared $L_2$ norm of the weights to the default cross-entropy loss function $\mathcal{L}$:

$$ \mathcal{L'} = \mathcal{L} - \delta \sum_{w \in \theta}|w|^2 $$

This pushes the weights to increase their norm. A simple manual search shows grokking to occur at $\delta = 2e-10$, while the weight norm keeps on increasing, as seen in Fig. \ref{fig:increasing_norm_mnist}. Note that for the AdamW optimizer, this is implemented differently than weight decay.

Thus, a high weight norm initialization and weight decay is not the cause for grokking as hypothesized by \citet{liu2022omnigrok}. In the next section, we motivate three progress measures that conceptually align with existing notions of generalization, and show that these measures better capture the occurrence of grokking in real-world datasets.

\section{Progress Measures for Grokking on Real-world Tasks}
\label{progress_measures}

\subsection{Activation Sparsity}

Let $\mathbf{a}_i$ be the activations of a given layer for input $x_i$. The sparsity $S$ is defined as the fraction of activations that are below a threshold $\tau$.

$$
S = \frac{1}{D} \sum_{i=1}^{D} \frac{1}{n} \sum_{j=1}^{n} \mathbb{I}(a_{i,j} < \tau)
$$

where $D$ is the number of training examples, $n$ is the number of activations in the layer, $\mathbb{I}(\cdot)$ is a condition indicator function, and $a_{i,j}$ is the $j$-th activation for the $i$-th input.

Activation sparsity measures how many activations are near zero, indicating how many neurons are effectively turned ``off'', with a higher sparsity implying that fewer neurons are actively contributing to the output.

Several prior works (such as \cite{peng2023theoretical, huesmann2021impact, li2022lazy}) study the properties of activation sparsity in the context of neural networks. Specifically, \citet{li2022lazy} find sparse MLP layers in networks that generalize well, \citet{merrill2023tale} develop a measure of \textit{effective sparsity} to explain grokking on an algorithmic task, and \citet{huesmann2021impact} observe an increase in activation sparsity right before an increasing test loss. In line with these observations, we observe activation sparsity to plateau right before the model starts to grok (see Figs. \ref{fig:decreasing_norm_mnist} and \ref{fig:increasing_norm_mnist}).

\subsection{Absolute Weight Entropy}

Let $\mathbf{W}$ be the weight matrix of a layer in a neural network with shape $(m, n)$. The Shannon entropy $H$ \citep{shannon2001mathematical} of the absolute weights is defined as:

$$
H(\mathbf{W}) = - \sum_{i=1}^{m} \sum_{j=1}^{n} |w_{ij}| \log |w_{ij}|
$$

While the weights of a network do not form a probability distribution, $abs(w_{ij})$ can be intuitively thought of as a measure of the impact of feature $i$ in layer $N$ affecting feature $j$ in layer $N+1$, where features are represented by neurons in the standard basis. While some model components represent features in superposition (such as a transformer's residual stream \citep{elhage2021mathematical}, MLP layers do have a privileged basis due to non-linearities \citep{elhage2021mathematical}. Thus, the absolute weight entropy captures the spread and magnitude of these weights. 

During training, the optimization process adjusts the weights to minimize the cross-entropy loss. Let's analyze how this adjustment impacts the absolute weight entropy.

\subsubsection{Initialization}

Typically, weights are initialized using schemes such as the Glorot uniform initialization \citep{kumar2017weight}, which distributes weights uniformly over an interval $[-\sqrt{\frac{6}{m+n}}, \sqrt{\frac{6}{m+n}}]$. This initialization is designed to keep the scale of gradients roughly consistent across layers. With $w_{ij} \sim U\left(a, a\right)$, where $a=\sqrt{\frac{6}{m+n}}$, the initial weight entropy is given by $ H(\mathbf{W}) = mn \left( \frac{a}{4} - \frac{a}{2} \log a \right) $. (See Appendix \ref{closed_form_proof} for a proof of the same).

\subsubsection{During Training}

During training, gradient descent iteratively updates the weights to reduce the cross-entropy loss $\mathcal{L}$. These updates are driven by the need to fit the training data better, which often leads to some weights increasing in magnitude (if they are critical for reducing the loss) and others decreasing or becoming very small (if they are less critical).

\textbf{Critical Weights Increase:} Weights that contribute significantly to reducing the loss may increase in magnitude. However, due to the lottery ticket hypothesis (shown to hold to a good degree for MLPs trained on MNIST \citep{frankle2018lottery}), this is often limited to a much smaller subset of the weights. Since \( |w_{ij}| \log |w_{ij}| \) increases with \( |w_{ij}| \), these weights contribute more to $H(W)$.

\textbf{Non-Critical Weights Decrease:} Many weights may decrease in magnitude as the model learns to focus on the most critical features. For weights close to zero, \( |w_{ij}| \log |w_{ij}| \) approaches zero, contributing less to $H(W)$.

Given these dynamics, one can expect that while a few weights might increase in magnitude, the overall trend for the majority of weights is towards smaller magnitudes, leading to $H(\mathbf{W}_{\text{trained}}) < H(\mathbf{W}_{\text{init}})$.
   
This reduction in entropy indicates a more concentrated weight distribution, which contributes to the model's robustness and generalization ability. Weight entropy measures the uncertainty or spread of the weight distribution, and a lower entropy is indicative of a compact and concentrated weight distribution, a sign of robust generalization.

We see in Figs. \ref{fig:decreasing_norm_mnist} and \ref{fig:increasing_norm_mnist} that the weight entropy sharply decreases with generalization in both increasing and decreasing weight norms and does not change during overfitting, verifying our hypothesis. Several prior works have studied different forms of entropy to regularize and stabilize neural network training \citep{gabrie2018entropy, zou2021increasing, araujo2022entropy} and to decrease their effective depth \citep{quetu2024simpler, wiedemann2019entropy}.

\subsection{Approximate Local Circuit Complexity}

Past works to calculate the effective circuit complexity of a neural network have been both computationally expensive and difficult to scale \citep{janik2020complexity, humayun2024deep}. \cite{humayun2024deep} create spline partitions on the network's input space to show that deep networks grok on adversarial datapoints. Here, we come up with a simple and scalable progress measure which we call the approximate local circuit complexity that is another meaningful measure of grokking (see Figs. \ref{fig:decreasing_norm_mnist} and \ref{fig:increasing_norm_mnist}).

Let $\mathbf{o}_h^{(\mathbf{W})}(x)$ be the output logits of a model for input $x$ with $\mathbf{W}$ being the weights of a layer of choice, and let $\mathbf{o}_h^{(\mathbf{W'})}(x)$ be the logits when 10\% of the model's weights are made zero, resulting in a new weight matrix $\mathbf{W'}$. Intuitively, this corresponds to turning off a random $10\%$ of the layer's computation in a similar fashion as dropout during training \citep{hinton2012improving}. The progress measure is then defined as the KL divergence $D_{\text{KL}}$ between the two logit distributions:

$$
D_{\text{KL}} \left( P(\mathbf{o}_h^{(\mathbf{W})} \| \mathbf{x}) \, \| \, P(\mathbf{o}_h^{(\mathbf{W'})} \| \mathbf{x}) \right) = \sum_{i=1}^{D} \sum_{y \in Y} P(\mathbf{o}_h^{(\mathbf{W})}(x_i, y)) \log \left( \frac{P(\mathbf{o}_h^{(\mathbf{W})}(x_i, y))}{P(\mathbf{o}_h^{(\mathbf{W'})}(x_i, y))} \right)
$$

where $P(\mathbf{o}_h^{(\mathbf{W})}(x_i, y))$ is the softmax probability distribution of logits under the original weights $\mathbf{W}$, and $P(\mathbf{o}_h^{(\mathbf{W'})}(x_i, y))$ is the same under perturbed weights $\mathbf{W'}$.

Conceptually, approximate local circuit complexity measures the sensitivity of the model's output to small perturbations in the weights. A lower KL divergence indicates that the model's logits are more robust to these perturbations, suggesting that the model has learned more stable and reliable internal representations, thereby leading to generalization. We empirically verify this hypothesis in Fig. \ref{fig:decreasing_norm_mnist}. We also show in Appendix \ref{dropout_mnist} that the same model trained with dropout does not show a delay in generalization.

Thus, we show that a decrease in the $L_2$ norm of the weights is not necessarily correlated with generalization as found by \cite{liu2022omnigrok}, and clearly doesn't imply a causation as claimed by their ``LU mechanism''. We show limited results on the IMDb dataset \citep{maas2011learning} in Appendix \ref{imdb}.

\section{Conclusion}

We introduce three conceptually motivated progress measures for grokking on real-world data and show that the correlation of grokking with a decrease in the $L_2$ weight norm is not required, and thus prove that weight norms are not the causal explanation behind grokking. We show generalization to occur at high weight norms while our measures still work well. As potential future directions, it would be interesting to see how the performance dynamics changes when we make these progress measure differentiable using approximations and add them to the loss function to control grokking. Another interesting direction would be to build a theoretical framework for grokking based on these measures and to concretely ascertain their occurrence during generalization. Lastly, it would be great to look at other real-world datasets to see the generalizability of the results.

\bibliography{main}
\newpage
\clearpage

\appendix
\section{Hyperparameters}
\label{hyperparams_mnist}

\begin{table}[h!]
\centering
\begin{tabular}{cc}
\toprule
\textbf{Hyperparameter} & \textbf{Value} \\
\midrule
Train Points & 1000 \\
Test Points & 1000 \\
Optimization Steps & 100000 \\
Batch Size & 1000 \\
Loss Function & MSELoss \\
Optimizer & AdamW \\
Weight Decay & 0.01 \\
Learning Rate & 1e-3 \\
Initialization Scale & 8.0 \\
Depth & 3 \\
Width & 200 \\
Activation & ReLU \\
\bottomrule
\end{tabular}
\caption{Hyperparameters used in the experiment. We train a 3-layer, 200-width MLP with ReLU non-linearities on the MNIST dataset using the AdamW optimizer with weight decay.}
\label{tab:hyperparameters}
\end{table}

\section{Closed-form Weight Entropy for Uniform Initialization}
\label{closed_form_proof}

Assume the weights are initialized uniformly over an interval \([-a, a]\). For a uniform distribution, each weight \(w_{ij}\) has an equal probability of taking any value within this interval. 

The closed-form for the weight entropy \(H\) can be derived as follows:

$$ w_{ij} \sim U(-a, a) $$
   
The weight log-entropy \(H\) is given by:
$$ H(\mathbf{W}) = - \sum_{i=1}^{m} \sum_{j=1}^{n} |w_{ij}| \log |w_{ij}| $$

For uniform initialization (such as the commonly used Xavier initialization scheme \citep{kumar2017weight}, let's assume that the weights are evenly distributed across the interval \([-a, a]\). We can approximate the sum by integrating over the interval for a continuous approximation:

$$ H(\mathbf{W}) \approx -mn \int_{-a}^{a} |w| \log |w| \cdot \frac{1}{2a} \, dw $$
   
Here, \(\frac{1}{2a}\) is the probability density function of the uniform distribution over \([-a, a]\). Splitting the integral into two parts (from \(-a\) to \(0\) and from \(0\) to \(a\)):

$$ H(\mathbf{W}) \approx -mn \left( \int_{-a}^{0} |w| \log |w| \cdot \frac{1}{2a} \, dw + \int_{0}^{a} |w| \log |w| \cdot \frac{1}{2a} \, dw \right) $$

Since \(|w| = -w\) for \(w < 0\) and \(|w| = w\) for \(w > 0\), we have:

$$ H(\mathbf{W}) \approx -mn \left( \int_{0}^{a} w \log w \cdot \frac{1}{a} \, dw \right) $$

This simplifies to:

$$ H(\mathbf{W}) \approx -mn \cdot \frac{1}{a} \int_{0}^{a} w \log w \, dw $$

To solve the integral, use integration by parts with \(u = \log w\) and \(dv = w \, dw\):

$$ \int_{0}^{a} w \log w \, dw = \left[ \frac{w^2}{2} \log w \right]_{0}^{a} - \int_{0}^{a} \frac{w^2}{2} \cdot \frac{1}{w} \, dw = \left[ \frac{w^2}{2} \log w \right]_{0}^{a} - \frac{1}{2} \int_{0}^{a} w \, dw $$
   
The second integral is straightforward:

$$ \int_{0}^{a} w \, dw = \left[ \frac{w^2}{2} \right]_{0}^{a} = \frac{a^2}{2} $$

Combining the results:

$$ \left[ \frac{w^2}{2} \log w \right]_{0}^{a} = \frac{a^2}{2} \log a - \left(0 \cdot \log 0 \text{ (which is 0)}\right) $$

$$ \int_{0}^{a} w \log w \, dw = \frac{a^2}{2} \log a - \frac{1}{2} \cdot \frac{a^2}{2} = \frac{a^2}{2} \log a - \frac{a^2}{4} $$

Therefore, we get the following closed-form expression for $H$:

$$ H(\mathbf{W}) \approx -mn \cdot \frac{1}{a} \left( \frac{a^2}{2} \log a - \frac{a^2}{4} \right) = -mn \left( \frac{a}{2} \log a - \frac{a}{4} \right) = mn \left( \frac{a}{4} - \frac{a}{2} \log a \right) $$

\section{Effects of Dropout}
\label{dropout_mnist}

We keep the rest of the setup the same as in Sec. \ref{progress_measures}, but add dropout \citep{hinton2012improving} to each linear layer of our neural network. We set the dropout fraction as $p=0.3$.

\begin{figure}
\centering
\includegraphics[width=\textwidth, trim=1.5cm 1.5cm 1.5cm 1cm, clip]{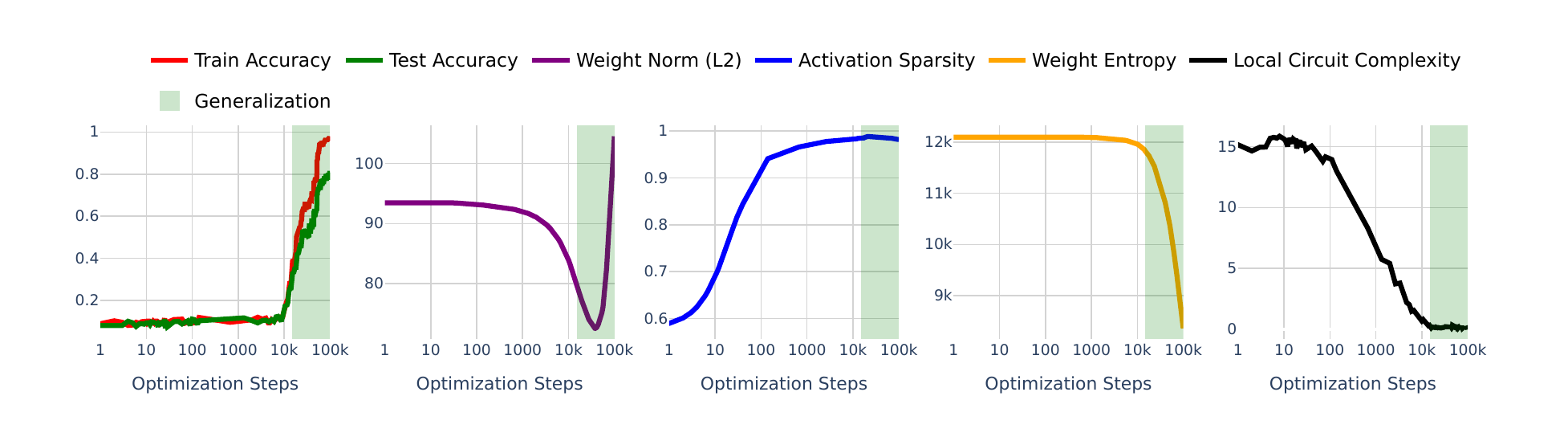}
\caption{The effects of dropout on delayed generalization and the progress measures. Note the absence of an overfitting zone as the model does not exhibit grokking.}
\label{fig:dropout_mnist}
\end{figure}

We see in Fig. \ref{fig:dropout_mnist} that the addition of dropout makes delayed generalization (grokking) completely vanish. This is in line with our experiments with the circuit complexity progress measure (see Sec. \ref{progress_measures}).

\section{Progress Measures on the IMDb Dataset}
\label{imdb}

\begin{figure}[h!]
\centering
\includegraphics[width=\textwidth, trim=1.5cm 1.5cm 1.5cm 1cm, clip]{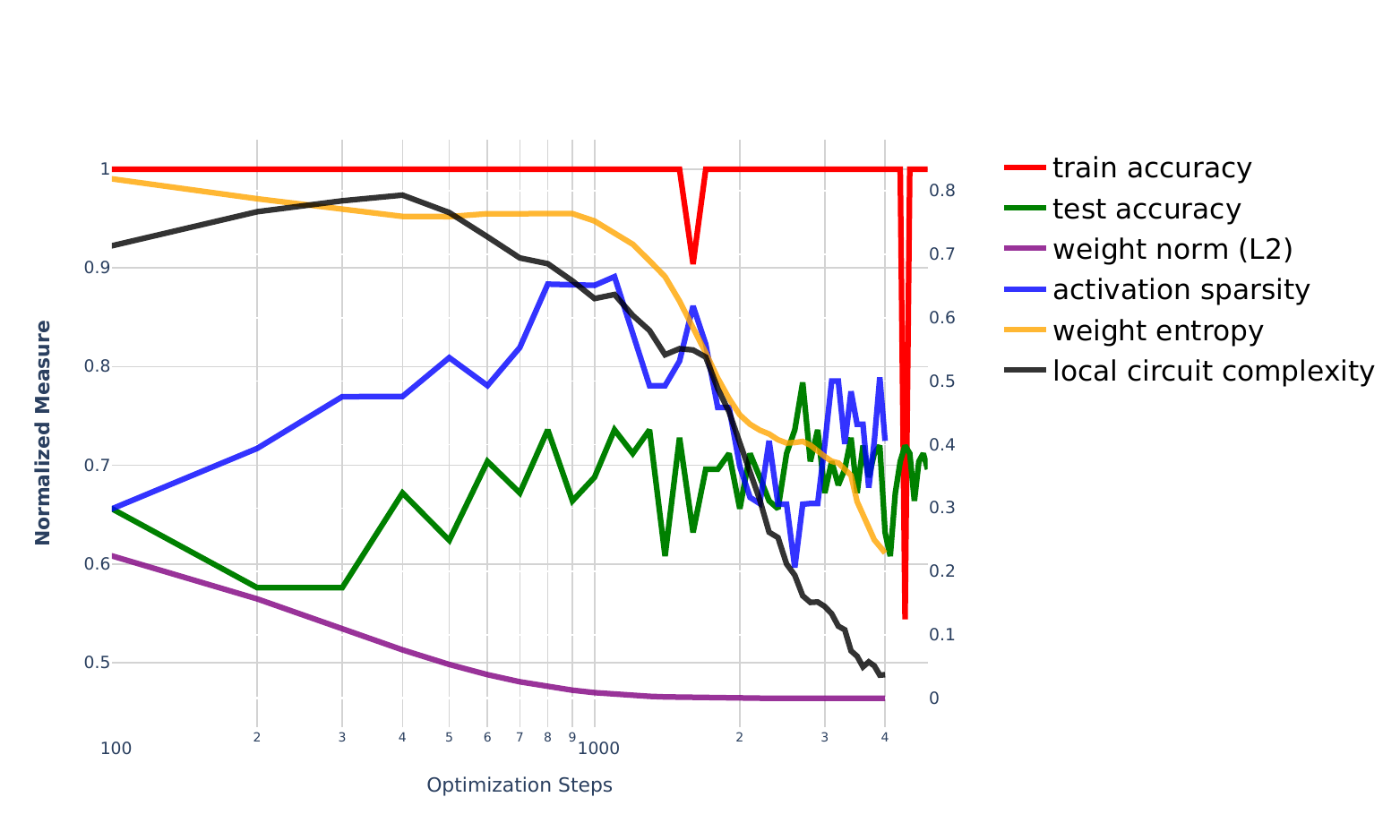}
\caption{Grokking, as observed on the IMDb dataset by \cite{liu2022omnigrok}, with a reduction in the $L_2$ norm of the weights and the new progress measures introduced.}
\label{fig:decreasing_norm_imdb}
\end{figure}

We show our progress measures on an LSTM-based \citep{vennerod2021long} model trained on the IMDb Movie Review Dataset \citep{maas2011learning}, with the exact same setup and hyperparameters as in \cite{liu2022omnigrok}. We observe in Fig. \ref{fig:decreasing_norm_imdb} that even with a different model architecture, our progress measures can be effectively used to study and predict generalization.

%\section{More}
%\input{appendix}

\end{document}